\documentclass{article}

	 \PassOptionsToPackage{numbers, compress}{natbib}
     \usepackage[final]{neurips_2020}

\usepackage[utf8]{inputenc} 
\usepackage[T1]{fontenc}    
\usepackage{hyperref}       
\usepackage{url}            
\usepackage{booktabs}       
\usepackage{amsfonts}       
\usepackage{nicefrac}       
\usepackage{microtype}      
\usepackage{amsmath}
\usepackage{amssymb}
\usepackage{graphicx}

\usepackage{caption}
\usepackage{subcaption}
\usepackage{multirow}
\usepackage{nicefrac}

\usepackage[usenames,dvipsnames]{xcolor}

\newcommand{\setA}{\mathcal{A}}
\newcommand{\setMetaA}{\mathcal{A}_{\mathit{meta}}}
\newcommand{\setI}{\mathcal{I}}
\newcommand{\reals}{\mathbb{R}}
\newcommand{\expectation}[1]{\mathbb{E}\left[#1\right]}
\newcommand*{\defeq}{\mathrel{\vcenter{\baselineskip0.5ex \lineskiplimit0pt
			\hbox{\footnotesize.}\hbox{\footnotesize.}}}%
	=}

\usepackage{ifthen}
\newboolean{blind}
\setboolean{blind}{false}

\title{Towards Meta-Algorithm Selection}

\ifthenelse{\boolean{blind}}
{
\author{
    Anonymous Author(s) \\
    \texttt{anon@ymous.net}
}}
{
\author{%
  Alexander Tornede, Marcel Wever, Eyke H{\"u}llermeier \\
  Heinz Nixdorf Institute and Department of Computer Science\\
  Paderborn University, Germany \\
  \texttt{\{alexander.tornede, marcel.wever, eyke\}@uni-paderborn.de} \\
}
}
\begin{document}

\maketitle

\begin{abstract}
  Instance-specific algorithm selection (AS) deals with the automatic selection of an algorithm from a fixed set of candidates most suitable for a specific instance of an algorithmic problem class, where ``suitability'' often refers to an algorithm's runtime.
  Over the past years, a plethora of algorithm selectors have been proposed.
  As an algorithm selector is again an algorithm solving a specific problem, the idea of algorithm selection could also be applied to AS algorithms, leading to a meta-AS approach: Given an instance, the goal is to select an algorithm selector, which is then used to select the actual algorithm for solving the problem instance.
  We elaborate on consequences of applying AS on a meta-level and identify possible problems.
  Empirically, we show that meta-algorithm-selection can indeed prove beneficial in some cases. In general, however, successful AS approaches have problems with solving the meta-level problem. 
\end{abstract}


\section{Introduction}
In instance-specific algorithm selection (AS), we are faced with instances of an algorithmic problem class, such as the Boolean satisfiability problem (SAT), and a set of algorithms to choose for solving the instance.
The goal of AS is to select the algorithm that is most suitable for a given instance, in the sense that it performs best with respect to a certain criterion.
AS is motivated by the observation that algorithms perform differently well on different instances, so that choosing in a per-instance manner can improve the overall performance \cite{survey_kerschkeHNT19}\,---\,quite in line with theoretical ``no free lunch'' results \cite{no_free_lunch_wolpert1997no}.

Several AS approaches have been proposed over the last years, the majority of which makes use of machine learning techniques in order to exploit properties of the instances for selecting a good algorithm. As an AS algorithm is again an algorithm solving a specific problem, the idea of algorithm selection could also be applied to AS algorithms, leading to a meta-AS approach. Indeed, a certain complementarity among AS algorithms can be observed (e.g.~\cite{tornede20_run2survive}) and the resulting meta-AS problem was first mentioned by \cite{lindauer2019algorithm} and \cite{survey_kerschkeHNT19}. Besides, the strategy of selecting among a set of meta-experts, who in turn select the expert actually solving the problem, appears to be quite common in practice. Consider, for example, the case of building a house.
Very few people select the handicraft companies actually building their house themselves, but rather choose a general contractor, who in turn picks craftsmen for building the requested house.

One of the few existing works is AutoFolio \cite{autofolio_lindauerHHS15}, suggesting combined selection and configuration of algorithm selectors.
However, it mainly focuses on the configuration of algorithm selection approaches on a scenario-level, i.e., it tries to find the best configuration of an algorithm selector for a collection of instances, also called scenario. Extending the idea of AutoFolio to the problem of automated machine learning, AS approaches optimizing design choices regarding automated machine learning strategies were proposed recently \cite{feurer2018towards,feurer2020auto}.
In this work, we seek to select among available algorithm selectors, which in turn select an actual algorithm solving the problem, \textit{for every instance individually}, allowing for much more nuanced decisions potentially yielding performance improvements.

\section{The (Per-Instance) Algorithm Selection Problem}
The per-instance algorithm selection problem, first introduced in \citep{algorithm_selection_problem_rice76}, comprises a problem instance space $\setI$ and a set of candidate algorithms $\setA$. The goal is to find a mapping
$s: \setI \longrightarrow \setA$, called algorithm selector.
Given a problem instance $i \in \setI$, the mapping should ideally return the algorithm $a^* \in \setA$ optimizing a performance measure $m: \setI \times \setA \longrightarrow \reals$.
This performance measure is usually costly to evaluate, rendering an enumeration over the set of algorithms infeasible.
Accordingly, the \textit{virtual best solver} (VBS) aka.~\textit{oracle}, i.e., the optimal algorithm selector, is defined as 
\begin{equation}
    s^*(i) = \arg\min_{a \in \setA} \expectation{m(i,a)}
\end{equation} for all problem instances $i \in \setI$, where the expectation accounts for the potential randomness in the application of an algorithm. The strategy always selecting the algorithm which is best on average, i.e., averaged over instances $i \in \setI$, is called the single best solver (SBS).

\subsection{Common Algorithm Selection Solutions}\label{sec:as_solutions}

To compensate for the costly to evaluate performance measure $m$, most AS approaches employ machine learning techniques for learning surrogate models of the form (or a similar one) $\widehat{m}: \setI  \times \setA \longrightarrow \reals$ approximating the original performance measure, while being fast to evaluate.
Such surrogates allow one to specify an algorithm selector $s: \setI \longrightarrow \setA$ as
\begin{equation}\label{eq:selector_based_on_estimate}
    s(i) \defeq \arg\min_{a \in \setA} \widehat{m}(i,a) \, .
\end{equation}
In order to learn such surrogate models, we assume the existence of training data in the form of a set of instances $\setI_D \subset \setI$ for which the performance $m(i,a)$ is available for some algorithms $a \in \setA$. In addition, we assume that instances can be represented as $d$-dimensional feature vectors using a feature function $f:\setI \longrightarrow \reals^d$. Examples of such features for an instance of the Boolean satisfiability problem (SAT) are the number of clauses or the number of variables. Note that the computation of these features costs time, which is important to consider when the measure $m$ to optimize is (related to) runtime. This means that algorithm selection approaches NOT requiring such features can be advantageous if the feature computation is time consuming.

A simple example approach inferring such a surrogate model in the form of a regression function was introduced in \cite{xu2007satzilla} (PAReg), where the authors suggest to learn a regression function $\widehat{m}_a: \setI \longrightarrow \reals$ for each algorithm $a \in \mathcal{A}$, predicting the performance of algorithm $a$ on a given instance. Hence a selector (\ref{eq:selector_based_on_estimate}) can be defined by $\widehat{m}(i,a) \defeq \widehat{m}_a(i)$. Another natural idea is to tackle the problem through multiclass-classification (MCC) methods where each algorithm corresponds to one class \cite{tornede20_run2survive}. Other approaches, such as SUNNY \cite{sunny_amadiniGM14} or ISAC \cite{isac_kadiogluMST10}, estimate the unknown performance of an instance/algorithm pair by falling back to similar instances, for which evaluations are available, using k-nearest neighbor or clustering techniques.
More recently, an effective decision-theoretic approach based on survival analysis was presented in \cite{tornede20_run2survive}, called Run2Survive (R2SPAR10 / R2SExp), explicitly modeling censored information present in the training data.
A more recent version of \cite{xu2007satzilla}, called SATzilla’11 \citep{satzilla11_xu2011hydra}, leverages pairwise comparisons of algorithms by training a cost-sensitive decision forest for each pair of algorithms, predicting which one most likely performs better for a given instance. The selected algorithm is determined via majority voting.
Following the idea of pairwise comparisons, \cite{hanselle2020hybrid} investigate how combined ranking and regression models perform for algorithm selection, demonstrating hybrid models to achieve performance superior to models considering only regression or only ranking information.
For a more comprehensive overview of algorithm selection, we refer to \cite{survey_kerschkeHNT19}.

\subsection{Loss Functions}
When considering algorithmic satisfaction problems, one is usually interested in optimizing the runtime as a performance measure $m$ under the condition that a given instance has to be solved.
However, in the case of combinatorial problems, some algorithms need extremely long to solve some instances \cite{gomes1997heavy}, resulting in a time-constrained evaluation.
For this purpose, one usually considers a so-called cutoff $C$, after which an algorithm is terminated and the instance considered unsolved.
The most common loss function in AS, taking this into account, is the penalized average runtime (PAR10)
\begin{equation}
    \text{PAR10}(t) =
    \begin{cases}
        t & \text{ if } t \leq C \\
        10 \cdot C & \text{ else }
    \end{cases}
\end{equation}
where $t$ is the runtime of the algorithm as chosen by the algorithm selector. This score is averaged over all instances in order to rate a selector's performance on a set of instances, also called scenario. As PAR10 scores are incomparable across scenarios, one often considers the normalized PAR10 score defined as $\text{nPAR10}(s) = (\text{PAR10}_{s} - \text{PAR10}_{\mathit{oracle}})/(\text{PAR10}_{\mathit{SBS}} - \text{PAR10}_{\mathit{oracle}})$, where $\text{PAR10}_{s}$ is the PAR10 score of an algorithm selector $s$.

\section{Per-Instance Meta-Algorithm-Selection}

Per-instance meta-algorithm-selection (meta-AS) concerns the problem of selecting an algorithm selector, for a given instance of an algorithmic problem class.
The selected algorithm selector then in turn selects the actual algorithm solving the problem. Hence, instead of directly choosing an algorithm to solve a problem instance, we take a detour by selecting an algorithm selector.

Formally, in the meta-AS problem, we seek to find a mapping $s_{\mathit{meta}}: \mathcal{I} \longrightarrow \setMetaA$, called algorithm selector selector (ASS), optimizing the original performance measure $m$.
While being in principle identical to the standard AS problem, the set of algorithms $\setMetaA$ to choose from does no longer contain algorithms actually solving problem instances, but algorithm selection approaches, which in turn choose one of the algorithms solving an instance. Hence, the set of meta-algorithms to choose from, is a subset of all possible algorithm selectors, i.e., $\setMetaA \subseteq \{ s \vert s: \setI \longrightarrow \setA \}$. 

\begin{figure}[t]
    \centering
    \includegraphics[width=0.9\linewidth]{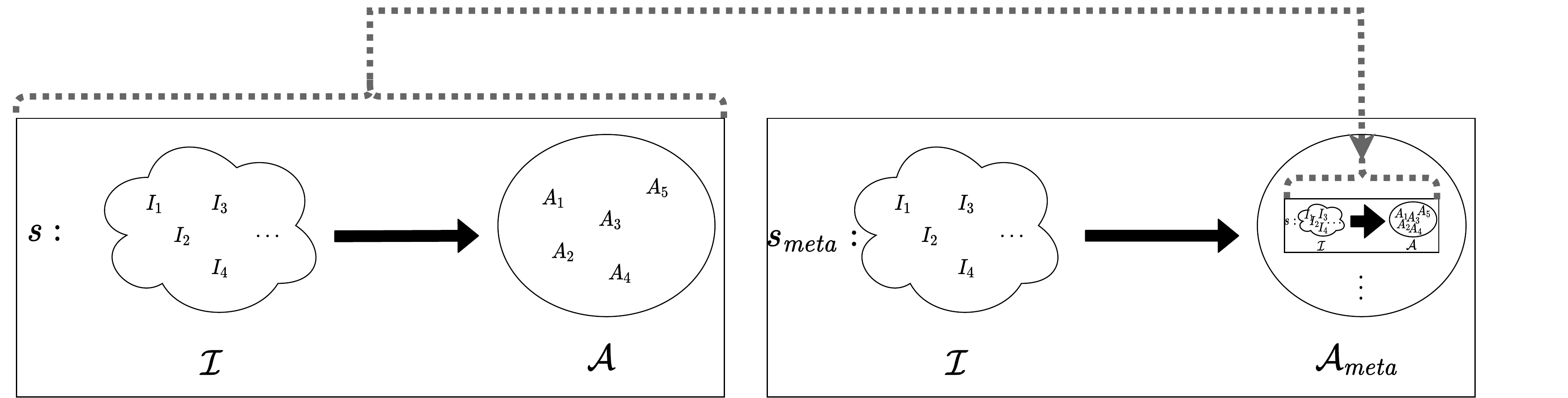}
    \caption{Left: An algorithm selector $s: \setI \longrightarrow \setA$ and its role as part of a meta-algorithm selector.
    Right: A meta-algorithm selector $s_{\mathit{meta}}$ that can choose from a set of algorithm selectors.}
    \label{fig:as_vs_meta_as}
\end{figure}

Similarly as in standard AS, we treat the algorithms, i.e., algorithm selectors, from $\setMetaA$ as black-boxes and ready to be applied, although one could in principle slot a training phase for the algorithm selectors in ahead. Hence, in the meta-algorithm selection problem one only has to derive a meta-algorithm selector. The connection between standard AS and meta-AS is depicted in Fig.~\ref{fig:as_vs_meta_as}.

\subsection{Meta-Algorithm-Selection Solutions}
As the meta-AS problem is essentially a special case of the standard AS problem with a very specific set of algorithms to choose from, standard AS methods (see Sec. \ref{sec:as_solutions}) can in principle be applied for solving the problem. It is important to note that algorithm selection approaches that do not rely on feature representations of instances do no longer necessarily have an advantage in terms of runtime, if they select an algorithm selector which in turn requires the feature representation. If the feature computation has to be performed either on the meta or on the base-level, its time has to be considered as well. However, there is no need to perform the computation twice if both the algorithm selector selector and the algorithm selector require it, as resulting features can be shared.

\subsection{Implications for the Oracle and the Single-Best-Solver}\label{sec:oracle_and_sbs_implications}
It is important to note that the oracle on the meta-level, i.e., the algorithm-selector-oracle (AS-oracle), is in general not identical to the oracle on the base-level, as the set of algorithms to choose from changes. In fact, the AS-oracle, is defined as
\begin{equation}
    s_{\mathit{meta}}^*(i) = \arg\min_{s \in \setMetaA} \expectation{m(i,s(i))} \,\,\, .
\end{equation}    
For a better understanding, consider the following example with two algorithms $A_1$ and $A_2$ and two algorithm selectors $s_1$ and $s_2$, where both always select algorithm $A_1$. Furthermore, assume that there exists an instance, for which $A_2$ performs better than $A_1$, and hence the oracle would select $A_2$. However, the AS-oracle can only select $s_1$ or $s_2$, which in turn both select $A_1$.

Generally speaking, in order to preserve the original oracle, there has to be, for each instance, at least one algorithm selector which selects the best algorithm. By the time this is not the case for at least one instance, the AS-oracle performance will degrade compared to the oracle. In practice, there will be at least one such instance most of the time, and hence an important question is how much the oracle performance degrades. As we show in our experimental evaluation, the degradation strongly depends on the scenario at hand, and ranges from less than $1\%$ to over $85\%$. 


Similarly to the oracle, the SBS changes as well, since the single best algorithm selector (SBAS), i.e., the SBS on the meta-level, is selecting from algorithm selectors, making it a lot stronger baseline. Hence, while the original oracle selects the actual problem solving algorithm best on average and accordingly does not depend on instance features, the SBAS does in fact depend on such features as long as it is not identical to the original SBS. Observe that this results in a significant disadvantage for the SBAS in terms of achievable PAR10 scores.

Obviously, these implications also influence the gains in performance, which can be achieved by algorithm selector selectors in comparison to algorithm selectors. As the oracle performance most likely degrades, while the SBS performance most likely improves, the gap between the two also decreases, offering less potential for algorithm selection approaches to close this gap.

\section{Experimental Evaluation}
In this section, we present the results of an experimental evaluation built around the meta-algorithm selection problem based on the ASlib v4.0 benchmark suite \cite{aslib_bischlKKLMFHHLT16}.
The study is organized into three parts:
First, we relate the performances of the SBS and the oracle for both levels to each other. Thereby, we discuss the fact that the gaps between SBS and oracle become smaller indeed, as already noted in Sec.~\ref{sec:oracle_and_sbs_implications}.
Second, we evaluate if and to what extent the application of algorithm selector selectors does yield an improvement in algorithm selection performance.
Third, we investigate whether algorithm selector selectors are able to improve over the SBAS.

We evaluate each approach for every ASlib scenario by performing a 10-fold cross-validation over the instances of a scenario and report cropped means for all measures we compute. For these cropped means, we remove the two best and two worst splits. Often algorithm selection systems consist of more elements than just a surrogate, such as pre-solvers or feature selectors. For our experimental study,
we consider pure algorithm selection models. All experiments were run on machines featuring Intel Xeon E5-2695v4@2.1GHz CPUs with 16 cores and 64GB RAM. In order to allow for full reproducibility of our results, the code for all experiments and the generation of plots including detailed documentation can be found on Github\footnote{\ifthenelse{\boolean{blind}}{Blinded for review}{\url{https://github.com/alexandertornede/meta_as}}}.

\subsection{Results}
\begin{figure}[t]
    \centering
    \includegraphics[width=0.9\linewidth]{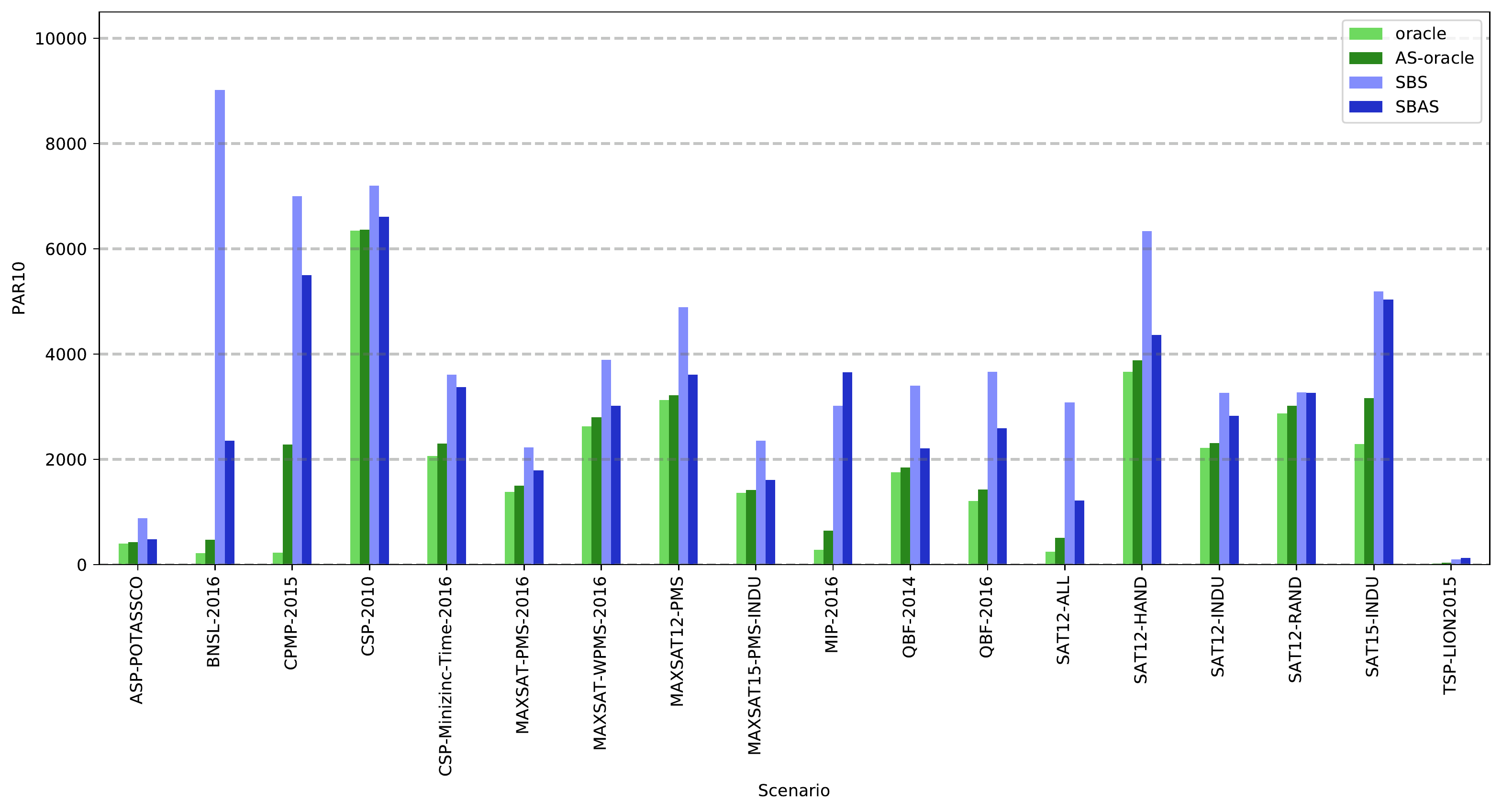}
    \caption{This figures shows the PAR10 scores of the oracle, AS-oracle, SBS and SBAS on a subset of the ASlib v4.0 benchmark scenarios as bar charts.}
    \label{fig:sbs_vbs_comparison}
\end{figure}

Fig. \ref{fig:sbs_vbs_comparison} shows the PAR10 scores of the oracle, AS-oracle, SBS and SBAS on a subset of the ASlib v4.0 benchmark scenarios. As one can see, several of the implications we noted in Sec. \ref{sec:oracle_and_sbs_implications} can be validated empirically: Firstly and most importantly, although the SBS/oracle gap is a lot larger than the SBAS/AS-oracle gap, the SBAS/AS-oracle gaps are non-negligible, and hence meta-algorithm selection does in principle make sense. For example, consider scenarios SAT12-ALL, or QBF-2016 with large SBAS/AS-oracle gaps.

As we noted earlier, the reason why these gaps become smaller is that oracle performance degrades when moving to the meta-level for all scenarios whereas the SBS performance improves when moving to the meta-level in general, since the SBAS is essentially an algorithm selector. While the degradation in oracle performance is moderate for the majority of scenarios (less than $10\%$), the SBS improvement is non-negligible, as the more successful the algorithm selectors considered by the algorithm selector selectors are, the larger this performance gain is.

\begin{table}[t]
    \centering
    \caption{Normalized PAR10 scores of all base- and meta-algorithm selectors normalized wrt. the standard oracle and SBS. The result of the best approach is marked in bold for each scenario. Moreover, for the meta-algorithm selectors the values in brackets $(a/b)$ indicate that that the approach achieves a performance better or equal to $a$ base-approaches and is worse than $b$ base-approaches.}
    \label{tbl:n_par10_level0}
    \resizebox{\textwidth}{!}{
        \begin{tabular}{l||rrrrrrr|rrrrrrr}
\toprule
\multicolumn{1}{c||}{Level} & \multicolumn{7}{c}{Algorithm Selectors} & \multicolumn{7}{|c}{Algorithm Selector Selectors (Meta)} \\
\toprule
\multicolumn{1}{r||}{\rotatebox{90}{Approach}} &  \multicolumn{1}{c}{\rotatebox{90}{R2SExp}} &  \multicolumn{1}{c}{\rotatebox{90}{R2SPAR10}} &   \rotatebox{90}{ISAC} &    \multicolumn{1}{c}{\rotatebox{90}{MCC}} & \multicolumn{1}{c}{\rotatebox{90}{PAReg}} &  \multicolumn{1}{c}{\rotatebox{90}{SATzilla'11}} &  \rotatebox{90}{SUNNY} &  \multicolumn{1}{c}{\rotatebox{90}{R2SExp}} &  \multicolumn{1}{c}{\rotatebox{90}{R2SPAR10}} &  \multicolumn{1}{c}{\rotatebox{90}{ISAC}} &  \multicolumn{1}{c}{\rotatebox{90}{MCC}} &  \multicolumn{1}{c}{\rotatebox{90}{PAReg}} &  \multicolumn{1}{c}{\rotatebox{90}{SATzilla'11}} &  \multicolumn{1}{c}{\rotatebox{90}{SUNNY}} \\
Scenario         &         &           &        &        &              &        &            &              &          &         &           &                 &           \\
\midrule

ASP-POTASSCO & 0.30 & 0.32 & 0.60 & 0.64 & 0.34 & 0.47 & \textbf{0.17} & 0.23 (6/1) & 0.22 (6/1) & 0.23 (6/1) & 0.37 (3/4) & 0.30 (6/1) & 0.30 (6/1) & 0.26 (6/1)\\
BNSL-2016 & \textbf{0.18} & 0.21 & 0.84 & 0.31 & \textbf{0.18} & \textbf{0.18} & 0.25 & 0.22 (3/4) & 0.21 (4/3) & 0.19 (4/3) & 0.28 (2/5) & 0.22 (3/4) & 0.28 (2/5) & 0.27 (2/5)\\
CPMP-2015 & 0.76 & \textbf{0.69} & 0.90 & 0.85 & 0.78 & 0.70 & 0.94 & 0.78 (4/3) & 0.78 (4/3) & 0.89 (2/5) & 0.81 (3/4) & 0.77 (4/3) & 0.81 (3/4) & 0.89 (2/5)\\
CSP-2010 & 0.13 & 0.15 & 0.31 & 0.80 & 0.25 & 0.13 & 0.34 & \textbf{0.04} (7/0) & \textbf{0.04} (7/0) & 0.19 (4/3) & 0.13 (7/0) & 0.46 (1/6) & 0.21 (4/3) & 0.09 (7/0)\\
CSP-Minizinc-Time-2016 & 0.43 & \textbf{0.27} & 0.83 & 0.36 & 0.67 & 0.34 & 0.37 & 0.51 (2/5) & 0.51 (2/5) & 0.76 (1/6) & 0.60 (2/5) & 0.67 (2/5) & 0.35 (5/2) & 0.51 (2/5)\\
GRAPHS-2015 & 0.76 & 0.71 & 0.88 & 1.37 & 0.80 & 1.02 & \textbf{0.45} & 0.76 (5/2) & 0.73 (5/2) & 1.06 (1/6) & 1.05 (1/6) & 0.87 (3/4) & 1.22 (1/6) & 1.00 (2/5)\\
MAXSAT-PMS-2016 & 0.60 & \textbf{0.36} & 0.82 & 1.06 & 0.77 & 0.62 & 0.41 & 0.64 (3/4) & 0.65 (3/4) & 0.60 (5/2) & 0.71 (3/4) & 0.82 (2/5) & 0.98 (1/6) & 0.75 (3/4)\\
MAXSAT-WPMS-2016 & 0.44 & 0.37 & 0.76 & 0.85 & 0.52 & 0.31 & \textbf{0.16} & 0.37 (5/2) & 0.39 (4/3) & 0.60 (2/5) & 0.60 (2/5) & 0.54 (2/5) & 0.43 (4/3) & 0.44 (4/3)\\
MAXSAT12-PMS & 0.22 & 0.23 & 0.47 & 0.40 & 0.28 & 0.24 & 0.29 & 0.25 (4/3) & 0.25 (4/3) & 0.20 (7/0) & \textbf{0.19} (7/0) & 0.32 (2/5) & 0.20 (7/0) & 0.21 (7/0)\\
MAXSAT15-PMS-INDU & 0.34 & 0.44 & 0.89 & 1.06 & 0.55 & 0.39 & \textbf{0.24} & 0.36 (5/2) & 0.57 (2/5) & 0.33 (6/1) & 0.39 (5/2) & 0.40 (4/3) & 0.51 (3/4) & 0.26 (6/1)\\
MIP-2016 & 1.07 & 0.96 & 0.76 & 1.13 & 1.58 & 1.05 & \textbf{0.68} & 1.00 (4/3) & 0.98 (4/3) & 1.07 (3/4) & 1.25 (1/6) & 1.48 (1/6) & 1.29 (1/6) & 0.77 (5/2)\\
PROTEUS-2014 & \textbf{0.41} & \textbf{0.41} & 0.64 & 0.84 & 0.45 & 0.58 & 0.47 & 0.46 (4/3) & 0.46 (4/3) & 0.48 (3/4) & 0.49 (3/4) & 0.53 (3/4) & 0.62 (2/5) & 0.53 (3/4)\\
QBF-2011 & 0.21 & 0.20 & 0.36 & 0.35 & 0.18 & \textbf{0.15} & 0.22 & 0.21 (4/3) & 0.21 (4/3) & 0.21 (4/3) & 0.21 (4/3) & 0.29 (2/5) & 0.25 (2/5) & 0.25 (2/5)\\
QBF-2014 & \textbf{0.26} & 0.28 & 0.52 & 0.59 & 0.32 & 0.31 & 0.31 & 0.30 (5/2) & 0.32 (3/4) & 0.32 (3/4) & 0.35 (2/5) & 0.41 (2/5) & 0.39 (2/5) & 0.35 (2/5)\\
QBF-2016 & 0.52 & 0.51 & 0.65 & 0.69 & 0.61 & 0.61 & \textbf{0.49} & 0.55 (4/3) & 0.55 (4/3) & 0.52 (5/2) & 0.53 (4/3) & 0.62 (2/5) & 0.57 (4/3) & 0.58 (4/3)\\
SAT03-16\_INDU & \textbf{0.71} & 0.76 & 0.98 & 0.99 & 0.77 & 0.82 & 0.82 & 0.92 (2/5) & 0.90 (2/5) & 0.78 (4/3) & 0.79 (4/3) & 0.81 (4/3) & 0.84 (2/5) & 0.86 (2/5)\\
SAT11-HAND & \textbf{0.34} & \textbf{0.34} & 0.65 & 0.57 & 0.46 & 0.44 & 0.60 & 0.42 (5/2) & 0.47 (3/4) & 0.42 (5/2) & 0.44 (5/2) & 0.50 (3/4) & 0.45 (4/3) & 0.56 (3/4)\\
SAT11-INDU & 0.69 & 0.69 & 1.08 & 0.71 & 0.63 & 0.79 & 0.76 & 0.78 (2/5) & 0.89 (1/6) & 0.84 (1/6) & \textbf{0.61} (7/0) & 0.79 (2/5) & 0.73 (3/4) & 0.85 (1/6)\\
SAT11-RAND & 0.13 & \textbf{0.06} & 0.58 & 0.17 & 0.09 & 0.39 & 0.12 & 0.13 (4/3) & 0.13 (4/3) & 0.19 (2/5) & 0.17 (3/4) & 0.19 (2/5) & 0.32 (2/5) & 0.20 (2/5)\\
SAT12-ALL & \textbf{0.36} & \textbf{0.36} & 0.67 & 0.38 & 0.37 & 0.44 & 0.38 & 0.37 (5/2) & 0.39 (2/5) & 0.39 (2/5) & 0.37 (5/2) & 0.40 (2/5) & 0.40 (2/5) & 0.43 (2/5)\\
SAT12-HAND & 0.34 & 0.34 & 0.65 & 0.41 & 0.37 & \textbf{0.27} & 0.43 & 0.34 (6/1) & 0.33 (6/1) & 0.31 (6/1) & 0.40 (3/4) & 0.40 (3/4) & 0.39 (3/4) & 0.38 (3/4)\\
SAT12-INDU & 0.70 & 0.73 & 1.02 & 0.94 & 0.79 & \textbf{0.59} & 0.78 & 0.62 (6/1) & 0.63 (6/1) & 0.75 (4/3) & 0.73 (5/2) & 0.65 (6/1) & 0.65 (6/1) & 0.66 (6/1)\\
SAT12-RAND & 0.96 & 0.86 & \textbf{0.84} & 5.20 & 1.17 & 0.93 & 1.14 & 1.03 (3/4) & 1.05 (3/4) & 0.95 (4/3) & 0.96 (4/3) & 1.19 (1/6) & 1.13 (3/4) & 1.05 (3/4)\\
SAT15-INDU & 0.95 & 0.83 & 0.76 & 0.91 & 0.74 & 0.75 & 1.00 & 0.68 (7/0) & 0.88 (3/4) & 1.00 (1/6) & 0.96 (1/6) & \textbf{0.65} (7/0) & 0.85 (3/4) & 0.81 (4/3)\\
TSP-LION2015 & \textbf{1.73} & 1.77 & 16.69 & 56.55 & 2.60 & 40.08 & 10.27 & 1.74 (6/1) & \textbf{1.73} (7/0) & 17.79 (2/5) & 36.21 (2/5) & 3.03 (4/3) & 44.3 (1/6) & 12.35 (3/4)\\
\bottomrule
\end{tabular}

    }
\end{table}

Table \ref{tbl:n_par10_level0} shows the normalized PAR10 scores of all algorithm selectors and the corresponding algorithm selector selectors, both normalized with respect to the standard oracle and SBS. The result of the best approach is marked in bold for each scenario. A value of 0 corresponds to oracle performance, a value of 1 to SBS performance, and values in-between indicate an improvement over the SBS, whereas values above 1 indicate degradation compared to the SBS. Unsurprisingly, most algorithm selector selectors are able to consistently improve over the SBS. However, moving to the meta-level proves to be beneficial for only 3 scenarios, which is disappointing. A possible cause of this observation might be that the considered AS approaches are not able to unleash their full potential on the meta-level, although considerable SBAS/AS-oracle gaps exist, as we have seen previously. In order to further investigate this, we evaluate the different meta-AS approaches w.r.t.\ their meta-algorithm selection performance in the following. 

Moreover, for the algorithm selector selectors, the values in brackets $(a/b)$ indicate that the approach achieves a performance better or equal to $a$ base approaches and is worse than $b$ base approaches. These results indicate that moving to the meta-level is beneficial in the sense that a more robust performance across several scenarios can be achieved. 

Table \ref{tbl:pairwise_comparison} shows a pairwise win/tie/loss between algorithm selectors and algorithm selector selectors aggregated across scenarios, giving a fairer idea of whether deciding for an algorithm selector selector is more beneficial than deciding for an algorithm selector. Although some ASS are able to beat many algorithm selectors on a majority of instances, ASS do not seem to be advantageous.

\begin{table}[t]
    \centering
    \caption{Win/Tie/Loss statistics, comparing ASS to algorithm selectors across scenarios.}
    \label{tbl:pairwise_comparison}
    \resizebox{.6\textwidth}{!}{
\begin{tabular}{ll||lllllll}
\toprule
&& \multicolumn{7}{c}{Algorithm Selector Selectors (Meta)}\\
&&R2SExp&R2SPAR10&ISAC&MCC&PAReg&Satzilla&Sunny\\
\midrule
\multirow{7}{*}{\rotatebox{90}{Alg. Selectors}}&R2SExp & 6/4/15 & 8/3/14 & 5/4/16 & 2/3/20 & 2/1/22 & 5/1/19 & 7/1/17\\
&R2SPAR10 & 6/2/17 & 5/1/19 & 5/0/20 & 4/1/20 & 4/0/21 & 3/0/22 & 7/0/18\\
&ISAC & 23/0/2 & 22/0/3 & 20/0/5 & 20/0/5 & 21/1/3 & 19/0/6 & 21/0/4\\
&MCC & 23/0/2 & 22/0/3 & 19/0/6 & 21/1/3 & 20/0/5 & 21/0/4 & 20/0/5\\
&PAReg & 17/2/6 & 13/2/10 & 11/1/13 & 11/1/13 & 6/1/18 & 11/0/14 & 11/0/14\\
&Satzilla & 14/0/11 & 9/0/16 & 11/0/14 & 9/3/13 & 8/1/16 & 6/0/19 & 11/0/14\\
&Sunny & 14/0/11 & 12/0/13 & 10/1/14 & 12/0/13 & 8/0/17 & 10/0/15 & 8/0/17\\
\bottomrule
\end{tabular}}
\end{table}

Table \ref{tbl:n_par10_level1} shows the PAR10 scores normalized w.r.t.\ to the SBAS and AS-oracle of all considered algorithm selector selectors, where improvements over the SBAS, i.e., values below 1, are marked in bold. A first observation to be made is that none of the meta-AS approaches is able to achieve an improvement over the SBAS on the majority of instances, indicating weaknesses in performing successful algorithm selection on the meta-level. This observation suggests that very successful AS approaches seem to have trouble to perform well on a meta-level. Note that these results have to be treated with care. Although the SBAS is essentially an algorithm selector, a score below 1 does not necessarily indicate that the corresponding ASS performs better than all algorithm selectors. The SBAS is determined on the training data, and hence an improvement over the SBAS on the test data is not equivalent to being better than \textit{all} algorithm selectors on the test data.

\begin{table}[t]
    \centering
    \caption{PAR10 scores normalized wrt. to the SBAS and AS-oracle of all considered meta-AS approaches. Values below 1, i.e., an improvement over the SBAS, are marked in bold.}
    \label{tbl:n_par10_level1}
    \resizebox{.65\textwidth}{!}{
        \begin{tabular}{l||rrrrrrr}
\toprule
Approach &  R2SExp &  R2SPAR10 &     ISAC &     MCC &    PAReg &  SATzilla'11 &   SUNNY \\
Scenario          &         &           &          &         &          &              &         \\
\midrule

ASP-POTASSCO & 3.15 & 2.03 & 1.46 & 5.46 & 2.77 & 2.80 & 1.71\\
BNSL-2016 & \textbf{0.87} & \textbf{0.74} & \textbf{0.86} & 1.41 & \textbf{0.97} & 1.17 & 1.20\\
CPMP-2015 & 1.05 & 1.05 & 1.17 & 1.33 & 1.12 & 1.01 & 1.22\\
CSP-2010 & 1.07 & \textbf{0.87} & 2.69 & \textbf{0.91} & 272.51 & 1.13 & \textbf{0.79}\\
CSP-Minizinc-Time-2016 & 1.64 & 1.64 & 2.04 & 14.69 & 3.11 & 1.03 & \textbf{0.94}\\
GRAPHS-2015 & \textbf{0.95} & \textbf{0.93} & 1.64 & 1.31 & 1.31 & 1.74 & 1.50\\
MAXSAT-PMS-2016 & 2.72 & 1.27 & 1.05 & 3.18 & 5.09 & 8.18 & 3.05\\
MAXSAT-WPMS-2016 & 1.20 & 5.44 & 3.84 & 11.1 & 7.83 & 4.44 & 3.37\\
MAXSAT12-PMS & \textbf{0.70} & \textbf{0.77} & \textbf{0.74} & \textbf{0.82} & 1.09 & \textbf{0.70} & \textbf{0.66}\\
MAXSAT15-PMS-INDU & 1.48 & 3.76 & 1.01 & 2.93 & 3.00 & 4.37 & 1.41\\
MIP-2016 & 1.01 & 1.00 & 1.13 & \textbf{0.99} & 1.00 & \textbf{0.95} & \textbf{0.76}\\
PROTEUS-2014 & 1.18 & 1.20 & 1.30 & 1.34 & 1.49 & 1.76 & 1.50\\
QBF-2011 & 1.34 & 1.49 & 1.67 & 1.22 & 1.85 & 1.75 & 2.04\\
QBF-2014 & 1.16 & 1.22 & 1.51 & 1.53 & 1.55 & 1.66 & 1.54\\
QBF-2016 & \textbf{0.91} & \textbf{0.96} & 1.00 & \textbf{0.88} & 1.09 & 1.20 & 1.09\\
SAT03-16\_INDU & 1.48 & 1.46 & 1.23 & 1.30 & 1.29 & 1.45 & 1.49\\
SAT11-HAND & 1.50 & 1.92 & 1.47 & 1.51 & 1.91 & 1.92 & 2.62\\
SAT11-INDU & 1.33 & 1.38 & 1.31 & \textbf{0.89} & 1.48 & 1.00 & 1.43\\
SAT11-RAND & 1.76 & 1.52 & 6.06 & 4.54 & 2.74 & 11.21 & 6.48\\
SAT12-ALL & 1.08 & 1.21 & 1.21 & 1.19 & 1.33 & 1.29 & 1.36\\
SAT12-HAND & 1.38 & 1.27 & 1.28 & 1.83 & 1.89 & 1.64 & 1.56\\
SAT12-INDU & 1.24 & 1.25 & 1.38 & 1.41 & 1.27 & 1.21 & 1.15\\
SAT12-RAND & 1.01 & 1.15 & 1.08 & 1.12 & 1.60 & 1.49 & 1.30\\
SAT15-INDU & \textbf{0.73} & \textbf{0.83} & 1.19 & 1.14 & \textbf{0.67} & \textbf{0.68} & \textbf{0.89}\\
TSP-LION2015 & 1.14 & 1.05 & 10.86 & 22.55 & 2.00 & 27.62 & 8.00\\
\bottomrule
\end{tabular}

    }
\end{table}

\section{Conclusion}

We proposed the problem of meta-algorithm selection, where the task is to select the right algorithm selector, and investigated how standard AS approaches perform on the meta-level.
Our analysis reveals interesting insights about the meta-AS problem concerning changes of the oracle and the SBS. As suggested by a considerable gap between the SBAS and the AS-oracle, a potential improvement through meta-AS definitely exists. This is confirmed by several cases, in which moving to the meta-level proves beneficial in terms of algorithm selection performance. In general, however, standard AS methods, when applied on the meta-level, do not seem to fully exploit this potential.

One may wonder, therefore, how AS on the meta-level could be improved. One idea is to consider features of the algorithm selectors that would allow the application of methods from extreme algorithm selection \cite{extreme_algorithm_selection_tornedeWH20,tornede2019algorithm}. Another idea worth investigation is to move to even higher meta-levels. Moreover, the problem of oracle degradation could be tackled by including the original algorithms (which one can think of as static algorithm selectors always choosing the same algorithm) in the set of algorithm selectors to choose from.


\newpage
\begin{ack}
This work was partially supported by the German Research Foundation (DFG) within the Collaborative Research Center ``On-The-Fly Computing'' (SFB 901/3 project no.\ 160364472). The authors gratefully acknowledge support of this project through computing time provided by the Paderborn Center for Parallel Computing (PC$^2$).
\end{ack}

\bibliographystyle{unsrt}
\bibliography{bibliography}

\end{document}